\documentclass[10pt,twocolumn,letterpaper]{article}

\usepackage{cvpr}
\usepackage{times}
\usepackage{epsfig}
\usepackage{graphicx}
\usepackage{amsmath}
\usepackage{amssymb}

\usepackage{multirow}

\usepackage{bbding}

\usepackage[pagebackref=true,breaklinks=true,letterpaper=true,colorlinks,bookmarks=false]{hyperref}

\cvprfinalcopy 

\def\cvprPaperID{729} 

\ifcvprfinal\fi
\begin{document}

\title{Improving object recognition performance by analyzing invariance \\ properties of deep neural networks}
\title{Evaluating object recognition accuracy and invariance using a large-scale controlled dataset}
\title{Enhancing accuracy and invariance of deep neural networks using a large scale controlled object dataset}
\title{Enhancing accuracy and invariance of deep neural networks using a large scale controlled object dataset}
\title{New insights into deep learning using a large scale controlled object dataset}
\title{What can we learn about CNNs from a large scale controlled object dataset?}

\author{Ali Borji\\
UCF\\
{\tt\small aborji@crcv.ucf.edu}
\and
Saeed Izadi\\
AUT\\
{\tt\small sizadi@aut.ac.ir}
\and
Laurent Itti\\
USC\\
{\tt\small itti@usc.edu}
}

\maketitle

\begin{abstract}
Tolerance to image variations (e.g. translation, scale, pose, illumination) is an important desired property of any object recognition system, be it human or machine. Moving towards increasingly bigger datasets has been trending in computer vision specially with the emergence of highly popular deep learning models. While being very useful for learning invariance to object inter- and intra-class shape variability, these large-scale wild datasets are not very useful for learning  invariance to other parameters forcing researchers to resort to other tricks for training a model. In this work, we introduce a large-scale synthetic dataset, which is freely and publicly available, and use it to answer several fundamental questions regarding invariance and selectivity properties of convolutional neural networks. Our dataset contains two parts: a) \textit{objects} shot on a turntable: 16 categories, 8 rotation angles, 11 cameras on a semi-circular arch, 5 lighting conditions, 3 focus levels, variety of backgrounds (23.4 per instance) generating 1320 images per instance (over 20 million images in total), and b) \textit{scenes}: in which a robot arm takes pictures of objects on a 1:160 scale scene. We study: 1) invariance and selectivity of different CNN layers, 2) knowledge transfer from one object category to another, 3) systematic or random sampling of images to build a train set, 4) domain adaptation from synthetic to natural scenes, and 5) order of knowledge delivery to CNNs. We also explore how our analyses can lead the field to develop more efficient CNNs. 

\end{abstract}

\vspace{-10pt}
\section{Introduction}

Object and scene recognition is arguably the most important problem in computer vision and while humans do it fast and almost effortlessly, machines still lag behind humans. In some cases where variability is relatively low (e.g., fingerprint or frontal face recognition) machines outperform humans but they don't perform quite as well when variety is high. Hence, the crux of the object recognition problem is tolerance to intra- and inter-class variability, lighting, scale, in-plane and in-depth rotation, background clutter, etc~\cite{dicarlo2012does}.

Thanks to deep neural networks, computer vision has enjoyed a rapid progress over the last couple of years witnessed by high accuracies over the ImageNet dataset (top-5 error rate about 5-10\% over 1000 object categories). These models (e.g.,VGG~\cite{SimonyanZ14a}, Alexnet~\cite{krizhevsky2012ImageNet}, Overfeat~\cite{sermanet2013overfeat}, and GoogLeNet~\cite{szegedy2014going}) have surpassed previous scores in several applications and benchmarks such as generic object and scene recognition~\cite{krizhevsky2012ImageNet,SimonyanZ14a}, object detection~\cite{sermanet2013overfeat,girshick2014rich}, semantic scene segmentation~\cite{chen2014semantic,girshick2014rich}, 
face detection and recognition~\cite{yang15from}, texture recognition~\cite{cimpoi15deep}, fine-grained recognition~\cite{lin15bilinear}, multi-view 3D shape recognition~\cite{su15multi}, activity and classification~\cite{simonyan2014two,karpathy2014large}, and saliency detection~\cite{kruthiventi2015deepfix}.

One big concern regarding the wild large scale benchmarks and datasets, however, is the lack of control over data collection procedures and deep comprehension of stimulus variety. While existing large-scale datasets are very rich in terms of inter- and intra-class variability, they fail to probe the ability of a model to solve the general invariance problem. In order words, natural image datasets (e.g., ImageNet~\cite{deng2009imagenet}, SUN~\cite{xiao2010sun}, PASCAL VOC~\cite{everingham2010pascal}, LabelMe~\cite{labelme}, Tiny~\cite{tiny}) are inherently biased in the sense that they do not offer all object variations. To remedy this, some works (e.g.,~\cite{Pinto}) have resorted to synthetic datasets where several object parameters exist.

Ideally, we want models to be tolerant to identity-preserving image variation (e.g. variation in position, scale, pose, illumination, occlusion). To probe this, some researchers have used synthetic home-brewed datasets either by taking pictures of objects on a turntable (e.g., NORB~\cite{norb}, COIL~\cite{coil}, SOIL-47~\cite{soil}, ALOI~\cite{GeusebroekIJCV2005}, GRAZ~\cite{graz}, BigBIRD~\cite{singh2014bigbird}) or by constructing 3D graphic models and rendering textures to them (e.g., Pinto~\cite{Pinto}, Saenko~\cite{peng2014exploring}). While being very beneficial in the past, these datasets are very small for training deep neural networks with millions of parameters. Further, they usually have small number of classes, instances per class, background variability, in plane and in-depth rotation, illuminations, scale, and total number of images. Here, to remedy these shortcomings, we introduce a large scale controlled object dataset with rich variety and a larger set of images.

Our main contributions in this work are two fold: 1) We introduce a large scale controlled dataset of objects shot in isolation and placed on scenes (together with other objects), and 2) We conduct several analyses of CNNs addressing fundamental questions and propose new pathways to build more efficient deep learning models in the future. 


\section{Related work}


Several controlled datasets for recognition tasks have been introduced in the past which have dramatically helped progress in computer vision. Some famous examples are FERET face~\cite{feret} and MNIST digit~\cite{lecun1998gradient} datasets. Nowadays, we have systems that perform either at the level of humans or superior (perhaps not as robust due to variations and noise). Similar datasets are available for generic object recognition but lack characteristics of a large scale representative dataset covering many sorts of invariance (e.g., background clutter, illumination, shape, occlusion, size). For example, the COIL
dataset~\cite{coil}, which also used a turntable to film 100 objects under various lightings and poses, only contains one object instance per category (e.g., one telephone, one mug). Further, objects were shot on only black backgrounds. As another example, the larger ALOI dataset~\cite{GeusebroekIJCV2005} contains 1,000 objects but few instances per category. The NORB dataset~\cite{norb} has 50 small toy objects (10 instances in each of 5 categories), however, all objects were painted uniformly and shot in greyscale on blank backgrounds. Almost all available turntable datasets are small scale and not very rich in terms of variations. 

Existing natural scene datasets such as ImageNet~\cite{deng2009imagenet}, SUN~\cite{xiao2010sun}, Caltect256~\cite{Griffin_etal07}, and Tiny~\cite{tiny} are very rich at the instance level but lack variety in terms of other parameters (e.g., many instances of an object such as car but only from a random viewpoint).

Most of previous research using controlled datasets, such as turntables images, has been focused on inspecting models or to brew concepts and ideas. Some recent works have attempted to show that there is a real benefit of these datasets and results achieved over them may generalize/transfer to large scale natural scene datasets. This has been studied under the names of domain adaptation or knowledge transfer. The idea here is that knowledge gained from a controlled dataset, created in one of the two ways mentioned above, can be transfered to real-world naturalistic datasets which may even have different statistics. For example, Peng et al.~\cite{peng2014exploring} trained a model from synthetically generated images (using a 3D graphics object model) and by augmenting their data with images from ImageNet and PASCAL, reported an improvement in accuracy over the latter datasets. They, however, did not probe whether what they learned was due to better invariance or richness at the instance level. Some other works have advocated and pursued this direction under different terminology ~\cite{gopalan2011domain,saenko2010adapting,draper2001adapting,fernando2013unsupervised}.

Another drive for using controlled datasets comes from neuroscience and cognitive vision literature. While CNNs were inspired by hierarchical structure of the human visual ventral stream~\cite{fukushima1980neocognitron}, they were later used to explain some physiological and behavioral data of humans and monkeys (e.g.,~\cite{hmax,serre,yamins2014performance,serre2007feedforward}). It has also been asserted that humans learn invariance with few presentations of an object a.k.a., zero- or one-shot learning. This is the opposite of the way that CNNs learn recognition. These models need an enormous amount of labeled data. In this work, we explore how a rich controlled dataset, containing a lot of information regarding various object parameters, can be utilized to improve object recognition performance. It is worth noting that being aware of human performance is important otherwise progress could get trapped in a local minima. Just recently He et. al.~\cite{he2015delving} reported top-5 error of 4.9\% over ImageNet which is lower than 5.1\% human error rate. This raises some questions such as: 
Have models surpassed humans? Is it theoretically possible to achieve a better performance than humans on this problems? etc.


Another related area to our work, which naturally fits well to turntable datasets, is the manifold embedding and dimensionality reduction literature. These techniques try to preserve and leverage the underlying low dimensional manifold in a supervised or unsupervised manner (e.g.,,~\cite{yuan2015scene,tomar2014manifold}). For instance, Weston et al.~\cite{weston2012deep} introduced an embedding-based regularizer to impose same labels for neighboring training samples thus benefiting from structure/manifold in the data. They used gradient descent to optimize the regularizer and adopted it for CNNs. Another classic example is Siamese Networks~\cite{bromley1993signature} which are two identical copies of the same network, with the same weights, fed into a `distance measuring' layer to compute whether the two examples are similar or not, given labeled data  which encourages similar examples to be close, and dissimilar ones to have a minimum certain distance from each other. While these techniques have been applied to controlled datasets, it still remains to explore how useful they are over large scale datasets. Our proposed dataset can be helpful in this direction as it combines the best of the two worlds: instance-level variety of large scale datasets and rich parametrized controlled synthetic images. These two, we believe, could be precious to enhance the capability of CNNs.

\begin{figure}[t]
\begin{center}
\includegraphics[width=1\linewidth]{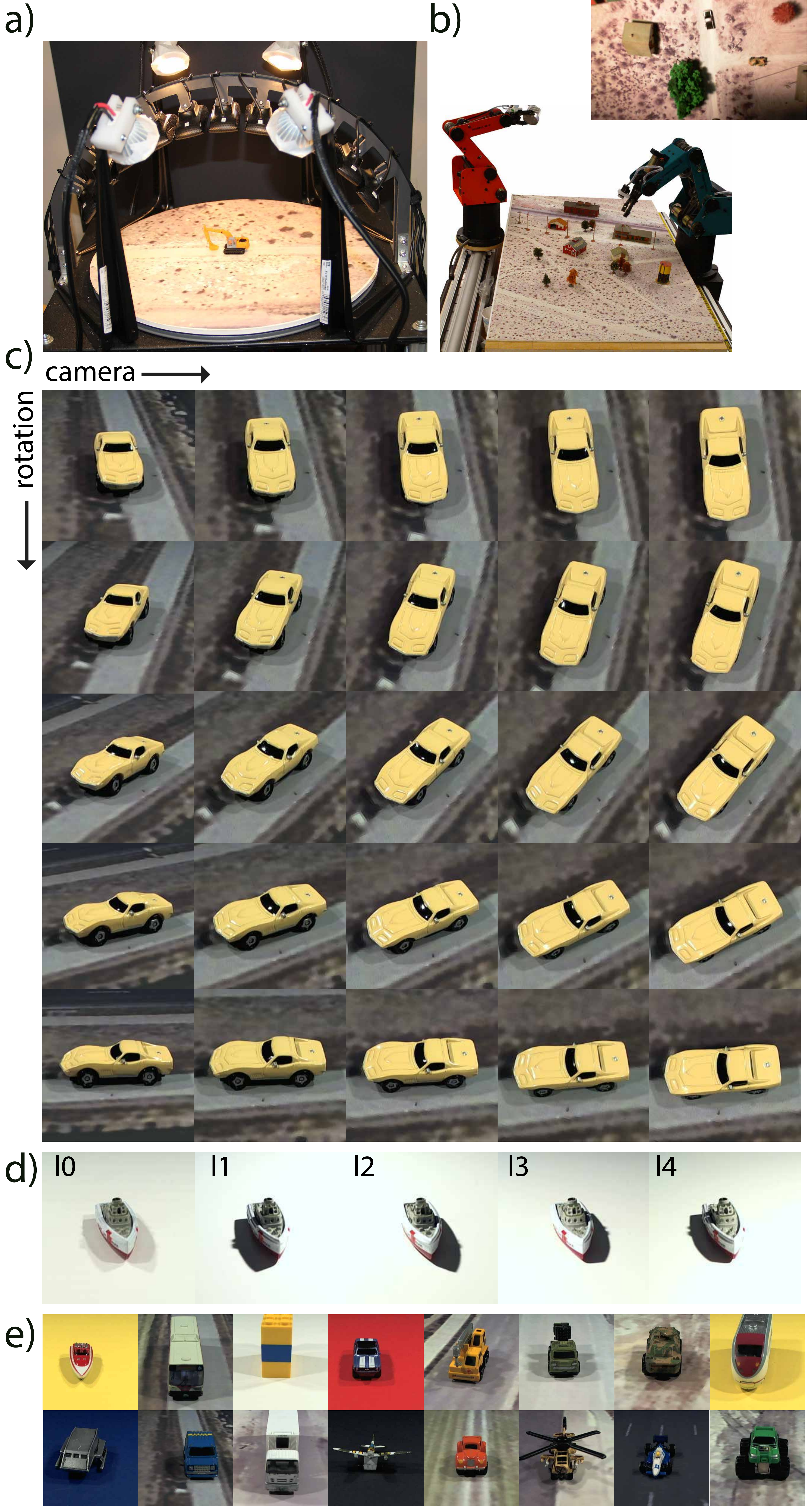}
\end{center}
\caption{a) \footnotesize{Turn-table object photo shooting setup. a) turntable with 8 rotation angles, 11 cameras on an arch, 4 lighting sources (generating 5 lighting conditions), 3 focus values and random backgrounds (overall 1320 images for each object instance per background). Recording parameters are: resolution 960 $\times$ 720, color mode YUYV, brightness 128, contrast
32, saturation 32, gain 30, auto white balance off, manual white balance temperature 3100K, sharpness
72, auto exposure off, auto focus off, focus base value 97-119. b) robotic-assisted arms, one holding camera, the other taking wide-field pictures from random viewpoints and distances. c) a sample instance of a car from 5 consecutive rotations and 5 consecutive arch cameras, d) an instance of a boat under different illuminations, and e) a sample instance from each object category (same lighting, rotation and focus all set to zero) presented in the order shown in Table~\ref{db}.}}
\vspace{-17pt}
\label{fig:dataset}
\end{figure}

\section{Turntable object dataset}

Our dataset contains 16 categories of objects(Micro machines toys produced by Galoob corp.) which differ in shape, texture, color, etc. It has 25-160 instances per category shot on about real world backgrounds (printed satellite images in the scale of 1:160). The whole dataset contains more than 20 million images and occupies about 17.65TB. We describe the photo shooting in the following. 

Each object instance was placed on a 14-inch diameter circular plate shown in Fig.~\ref{fig:dataset}.a. Turntable rotated 45 degrees per move thus generating 8 images (azimuth angles). This is referred to as `rotation' parameter in the rest of the paper. Eleven cameras (Logitech C910 webcams) were mounted on a semi-circular arch capturing 11 in-depth rotation images (elevation angles, referred to as `camera' parameter). We had four light sources (LED lightbulbs by Ecosmart ECS) placed on four corners of the table generating 4 lighting conditions (plus an additional fifth case where all lights were on). We also had 3 scales/focus conditions (-3, 0, and +3 from the default focus value of each camera). This setting resulted in a total of 1,320 images for one object instance on each background\footnote{Background scenes were 125 satellite imagery, randomly taken from the Internet, plus additional 7 plain backgrounds (white, red, blue, yellow, etc). Every object was photographed on at least 20 different related context backgrounds (e.g., boats on the water, cars on roads).} (11$\times$8$\times$5$\times$3). Images are in the color format with resolution\footnote{Cropped versions in 256 $\times$ 256 pixels are also available.} of 960 $\times $ 720 and are stored in the lossless PNG format (about 1MB each). Sample images of the dataset are shown in Fig.~\ref{fig:dataset}.c (rotations and camera images of an instance from the car category), Fig.~\ref{fig:dataset}.d (an instance of a boat shot under 5 different illuminations), and Fig.~\ref{fig:dataset}.e (samples of each object at rotation 0, lighting 0, focus 0 on a random chosen background). Statistics of the dataset are summarized in Table~\ref{db}.

As part of our dataset, we also shot objects in scenes using two robotics-assisted arms (Fig.~\ref{fig:dataset}.b). Several objects were placed manually on a congruent background (1:160 satellite maps corresponding to a 195m $\times$ 118m field).
One robot arm held a light source while another one carried the camera. Robots were programmed to: a) move randomly (flyby mode) and capture images in random positions, or b) target an object at a specific location and capture several images from different angles and distances. Images captured in this way have the resolution of 1280 $\times$ 720 pixels. While the turn table images support learning object recognition, robotics-assisted scenes offer a platform for training object detectors and scene understanding. Turntable images are from predefined parameters while images using the robotics workspace contain higher variety in terms of parameters (e.g., random viewpoints or scales). Together, these two types of images can be very useful for training and testing object detection and recognition models in a way that resemble natural settings.

\begin{table*}
\renewcommand{\tabcolsep}{.9mm}
\renewcommand\arraystretch{1.5}
\begin{center}{\footnotesize
\begin{tabular}{l|rrrrrrrrrrrrrrrr}
Category  & boat & bus & calib- & car  & equip- & lightweight & tank & train & ufo & van & semi  & air  & pickup  & heli-  & f1-car &  monster \\

  &  & & ration &   & ment & military &  & wagon &  &  & truck  & plane  & truck  & copter  &  &  truck \\

\hline


Num objects  & 27 & 25 & 13 &  160 & 64 & 54 & 31 & 25 & 40 & 29 & 33  & 85  & 40  & 25  & 40 & 40  \\


Num bg (mean) &  20 &21.3 &1 &26.1 &21.6 &18.5 &30.3 &37 &29 &29.4 &23.1 &18.4 &30.1 &23.2 &14 &21.5 \\

Num bg (std) &  0.0 &1.5 &0.0 &1.3 &1.3 &0.9 &7.8 &0.0 &4.4 &0.9 &5.0 &3.3 &4.9 &10.6 &0.0 &4.8 \\

Num bg (min-max) &  20-20 &20-23 &1-1 &24-28 &20-23 &18-20 &20-36 &37-37 &26-37 &28-30 &17-27 &17-26 &25-35 &14-35 &14-14 &14-25 \\

Total images  & 713K & 704K & 17K &  5517K & 1822K & 2611K  & 1432K & 462K &  739K&  933K &  1112K & 1907K   &   1505K& 660K& 950K & 1425K  \\


Size (GB)  & 551  &  545 & 11 & 4300  & 1500 & 2100 & 1200 & 363 & 565 & 724  & 874  &  1400 & 1200  &  495 & 722 &  1100\\


Used here  & $\checkmark$ & $\checkmark$ & - & -  & - & - & $\checkmark$ & $\checkmark$  & $\checkmark$  & $\checkmark$ &  - &  - &  - & -  & $\checkmark$ & -  \\

\end{tabular}}
\end{center}
\caption{Summary statistics of our dataset. There are 22,510,168 images in total from 16 categories (one used for calibration purposes only) with 25 to 160 instances per category. Five parameters include: 11 cameras on an arch, 4 lighting sources on 4 corners (5 conditions), 8 horizontal turn table rotations, 132 backgrounds (7 solid color) and 3 focus values. Average number of  backgrounds per object instance is 23.39. There are 46 unique backgrounds per category (avg bg per object 145.76 with std = 162.62; min = 25, max = 731). Total size of the dataset with resolution  960 $\times $ 720 is 17.65T. The cropped version of these images (256 $\times$ 256 pixels) is also available with 2.2TB in size. Total number of images per category is rounded to save space.}
\label{db}
\vspace{-12pt}
\end{table*}


\section{Results}

To start exercising the dataset, we tested it on small subsets of the available data. To understand generalization across image variations (object shape, object viewpoint, lighting, etc), CNNs are evaluated by taking slices of the dataset. We utilize a deep CNN pre-trained on ImageNet and fine-tune it on our dataset. The behavior of off-the-shelf features is investigated in our analyses as well. We use 7 object categories (out of 16) and avoid data augmentation as we have flipped versions of the objects from the turntable.

Since Alexnet has achieved great success in object and scene classification benchmarks, we choose it as the representative of CNNs in our analyses. Alexnet architecture is basically a linear feed-forward cascade of convolution and pooling layers as follows: the first two layers are
composed of 4 sublayers: \textit{convolution, local response normalization, ReLUs and max-pooling}.
Layers 3 and 4 include \textit{convolution and ReLUs} followed by Layer 5 which consists of \textit{convolution, ReLUs and max-pooling}. Two fully connected layers (fc6 and fc7) are then appended on top of the pool5 layer.
Finally, the fc8 is the label layer. We refer the reader to the original paper of Alexnet~\cite{krizhevsky2012ImageNet} for more details on model parameters (e.g., data augmentation, RGB jitter, etc). Depending on our analyses here the label layer may contain 2, 4 or 7 units (for object categories) or variable number of units depending on the parameter prediction task. We report average accuracies and standard deviations where there is randomness in the experimental procedure. Experiments are performed using the publicly available Caffe toolkit~\cite{jia2014caffe} ran on a Nvidia Titan X GPU and Ubuntu 14.04 OS.

We aim to answer these questions systematically: Can a pre-trained CNN model predict the setting parameters, say lighting source, degree of azimuthal rotation, degree of camera elevation, etc? and transfer the learned knowledge from one object category to another? Which parameters are more important in the transfer? How much knowledge can a model transfer from our dataset to the ImageNet dataset? What is a good strategy to make an object dataset? random or systematic image harvesting? and finally how the order of learning parameters invariance influences overall network parameter tolerance?

\subsection{Selectivity vs. invariance}

Humans are very good at predicting the category of an object and also tell about its setting parameters. 
This makes them selective (to parameters including object category) and invariant to variations.
In this experiment, we aim to systematically investigate this competition for two layers of the Alexnet: pool5 and fc7. We probe the expressive power of these layers for object and parameter prediction.

\begin{figure}[b]
\begin{center}
   \includegraphics[width=1\linewidth]{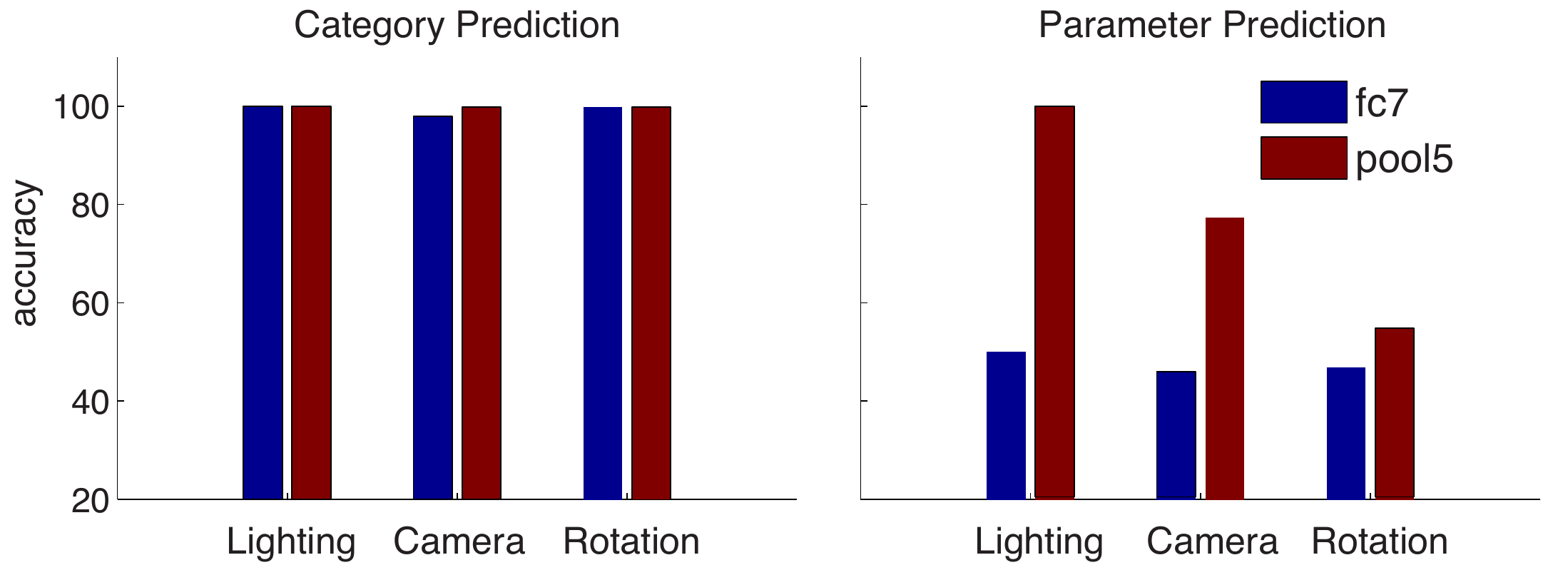}
\end{center}
\vspace{-10pt}
\caption{ِAnalysis of selectivy vs. invariance (expressive power) of pool5 and fc7 layers of Alexnet for category and parameter prediction over a four class problem.}
\label{fig:selvsInv}
\end{figure}

Four categories from our dataset (out of 16) were chosen for this analysis including boat, bus, tank and ufo. Images were lumped to train a SVM classifier. All features were normalized to have zero-mean before feeding to the classifier. The dimensionality was reduced to N-dimensions using SVD, where N refers to the number of instances in the training set. The reported results are average accuracy over random 5-fold cross validation test sets, each of size 2K. We trained two SVMs, one for category prediction and one for parameter prediction. Results are shown in Fig.~\ref{fig:selvsInv}.

As expected, we can see that fc7 features result in a high accuracy in classification, however, the surprising salient result is the shoulder-to-shoulder performance of pool5 to fc7. Relying on this outcome, it is clear that both fc7 and pool5 representations convey useful discriminative statistics for object recognition. Comparing the performance over parameter prediction, one can notice the superiority of pool5 layer over fc7. This is consistent with the work by Bakry et al.~\cite{bakry2015digging} where they analytically find that fully connected layers make effort to collapse the low-dimensional intrinsic parameter manifolds to achieve invariant representations. However, in Bakry et al.'s work, only view-manifold has been taken into consideration, while here thanks to our dataset, we can analyze the behavior of more common parameters in the real world.

In brief, it is clear that the feature space by pool5 contains much more knowledge than fc7 for parameter prediction. At the same time, the very representation makes different categories to be highly separable from each other (i.e., keeping the structure of manifolds as linearly-separable as possible for different categories). The representation by fc7 sensibly throws away the parameter information to become invariant while keeping the categories as separable as possible. We observe that the layer just before fully connected ones provides better compromise between categorization and parameter estimation.  

Parameter prediction accuracies for lighting, rotation, and camera view in order are 100\%, 77\%, and 62\%. This demonstrates that camera view has the most complex structure for parameter prediction whereas the lighting has the simplest. This is acceptable since changing camera view leads to geometric variations in the shape of the object, and ports the prediction task into a much more difficult problem to address. In contrast, lighting variations do not alter the shape of the object, and are thus easy to capture.

\begin{figure}[t]
\begin{center}
   \includegraphics[width=1\linewidth]{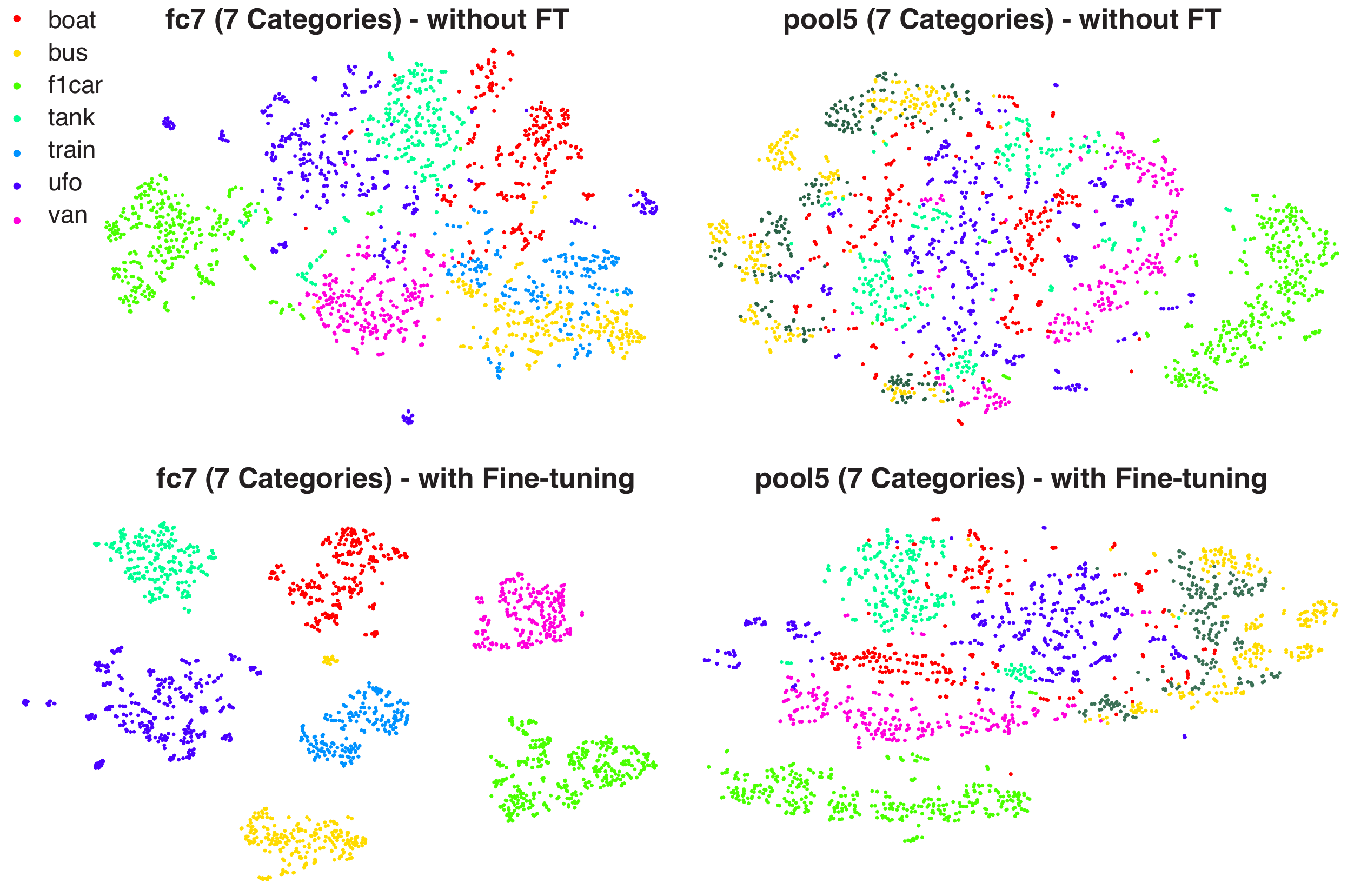}
\end{center}
\vspace{-5pt}
   \caption{t-SNE representation of Alexnet. The fc7 representation works remarkably well at recognizing object categories as they are mutually linearly separable after fine-tuning. Further, pool5 representation does not contain discriminative information compared to fc7. This figure also demonstrates the effect of fine-tuning. The distributions of samples for different categories tend to become very compact after fine-tuning. Notice that fine-tuning does not add more discriminative power to the pool5 representation.}
\label{fig:category}
\vspace{-13pt}
\end{figure}

We use the t-SNE dimensionality reduction method~\cite{van2008visualizing} to visualize the learned representations over seven categories from our dataset along with variation parameters (See Figure\ref{fig:category}). Please see also the supplement for details.

\subsection{Knowledge transfer}

Humans are very efficient to estimate and transfer parameters of a seen object to another object under many complicated scenarios. For example, they can reliably estimate the lighting source of an object and tell whether another object has been shot under nearly the same source direction. Complementary to our previous analysis, in this experiment, we aim to asses the power of CNNs in transferring the learned parameter over one object category to another. We focus on pool5 layer here since as we discussed, fc7 is invariant to parameters and not useful for discriminating between different values of parameters. 



\begin{figure}[t]
\begin{center}
   \includegraphics[width=1\linewidth]{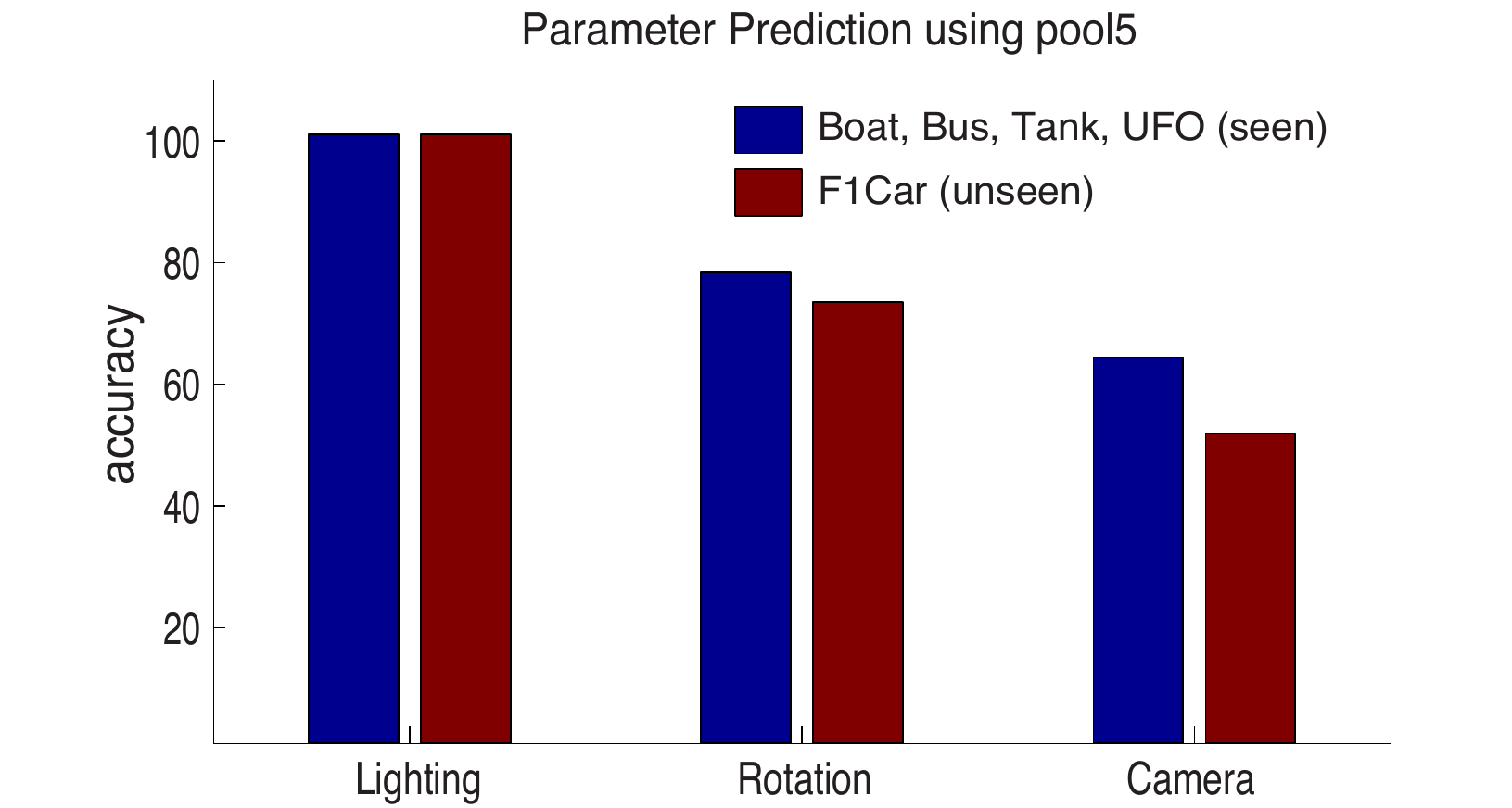}
\end{center}
\vspace{-5pt}
\caption{Knowledge transfer over different objects categories with one parameter changing for Alexnet trained over four classes and tested on same classes (but different instances) and f1car.}
\label{fig:knowledge}
\label{fig:onecol}
\vspace{-15pt}
\end{figure}

All parameters are fixed except one (i.e., slicing the dataset along only one parameter). 
We include instances from four categories (boat, bus, tank, ufo) in the training set, and test the learned knowledge on instances from an unseen category (f1car- red bar) as well as four seen categories (blue bar). We utilize the pool5 representation and reduce the dimensionality to N, where N refers to number of samples. The 5-fold cross validation average accuracy for parameter prediction is shown in Fig.~\ref{fig:knowledge}.

Results show a descent amount of knowledge transfer. It is observable from Fig.~\ref{fig:knowledge} that lighting parameter has the simplest knowledge to be transferred on unseen categories, as it has a head-to-head accuracy across seen and unseen categories. On the other hand, knowledge transfer for rotation and camera view is accompanied with sensible degradation in performance. To sum up, we see that the knowledge is promisingly transferable across seen and unseen categories, while the degradation in rotation and camera prediction is intuitively justifiable, as learned statistics in rotation and camera view prediction are dependent on the 3D properties of the object shape. 


%


\subsection{Systematic vs. random sampling}

Currently, large scale datasets are constructed by harvesting images randomly from the web. The main reason to do this is to include as much variability as possible in the dataset (mainly sampled along the intra- and inter-class variation).
While reasonable, it has not been systematically studied whether this is a good strategy compared to controlled ways conducted in turntable datasets. In this analysis, we consider two strategies to find the answer\footnote{We have a fixed test set from our dataset, and investigate which sampling strategy works better on this set.}: 1) Random strategy where we choose \textit{n} random samples (across all parameters and instances) and train an SVM to predict the object category, and 2) Systematic (or exhaustive) strategy, in which we choose an object instance randomly and then add images to our training pool, by scanning all parameters, until we reach \textit{n} samples. Assumption is that a fixed budget (time or cost) for processing only \textit{n} images is available.

\begin{figure}[t]
\begin{center}
   \includegraphics[width=1\linewidth]{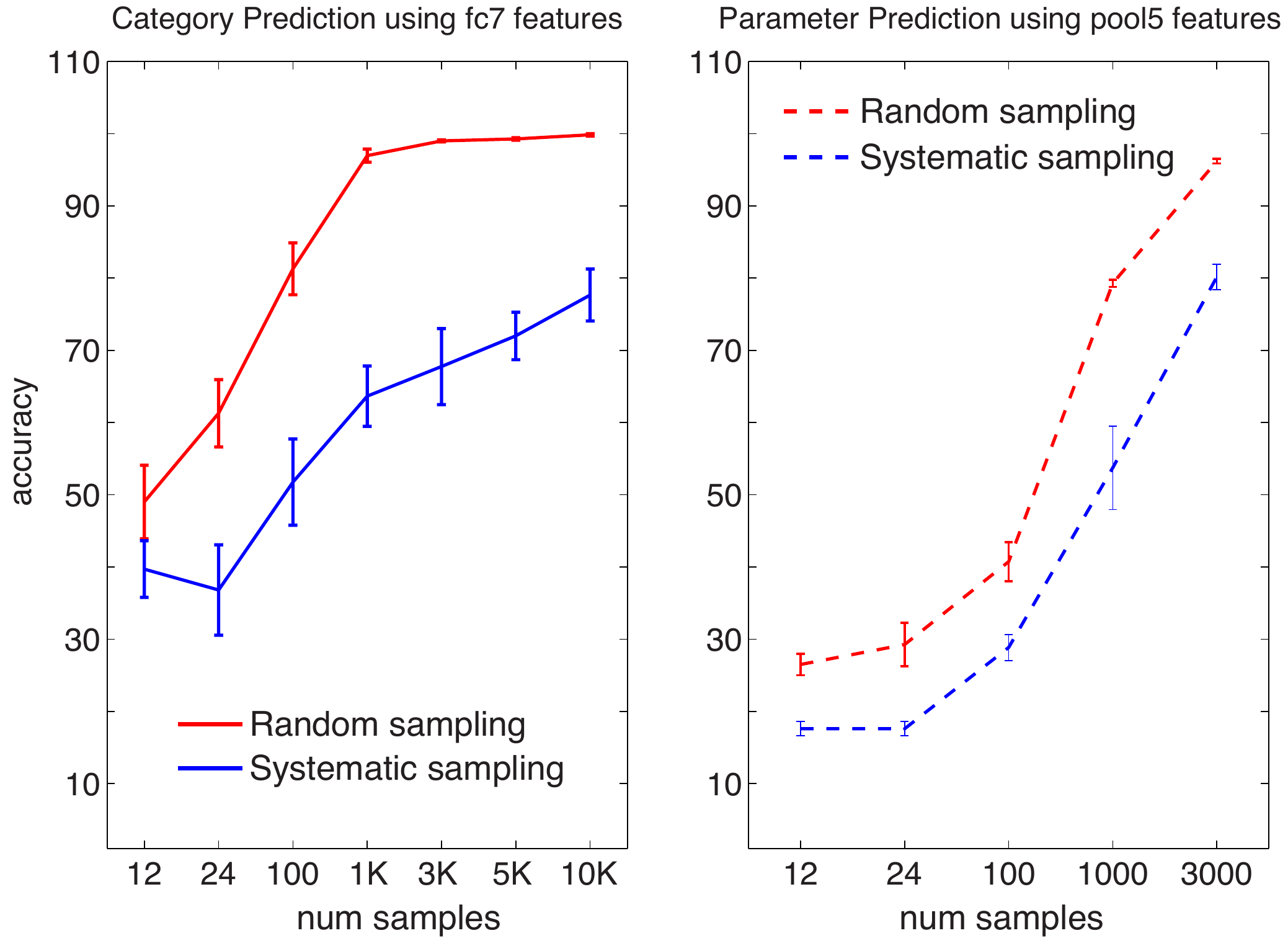}
\end{center}
\vspace{-5pt}
   \caption{Analysis of sampling strategies over a 4-class problem (boat, bus, tank, ufo). Left: category prediction accuracy using fc7 features. Right: Parameter prediction accuracy.}
\label{fig:sampling}
\label{fig:onecol}
\vspace{-10pt}
\end{figure}

We addressed a 4 class problem (boat, bus, tank, ufo) by increasing n starting from 12 up to 10000 samples. In each experiment, n/4 samples were chosen randomly from all 4 categories across all parameters, and were fed into the AlexNet to get the fc7(or pool5) representation. Then, we trained a linear SVM classifier on this data. A fixed test set of size 500 was randomly selected from all categories with all parameters and was kept fixed during the analysis. We measure category prediction at fc7 and parameter prediction at pool5, reducing the dimensionality to 2500 for all values of \textit{n} in the latter. Results are shown in Fig.~\ref{fig:sampling}.

We observe that random sampling strategy performs better in category prediction. This makes sense since randomly choosing images offers more instance level variety (better than systematic) leading to better recognition. Interestingly, and counter-intuitively, we see that random strategy works better in parameter prediction as well. We believe that the parameter prediction is somewhat dependent on the 3D properties of object shape, and since in the systematic strategy, the learner is not faced with sufficient instances, it fails to predict parameters compared to random strategy. Overall, what we learn is that instance level variations is of high importance in both category and parameter prediction and this is perhaps why the systematic sampling strategy is hindered. Thus, in dataset creation, it is vitally advantageous to have as much instance level variation as possible.

\subsection{Domain adaptation}

Currently, there is a gap in the literature connecting results learned over synthetic datasets with results on large scale datasets. One way that we pursue here is training models on our dataset (source) and see how much knowledge those models can transfer to the large scale wild datasets (target). This way, we discover along which dimension(s) a wild dataset varies the most and whether the target dataset offers sufficient variability for learning invariance. In other words, we can somehow indirectly measure dataset bias. Ultimately, we would like to generalize what we learn from synthetic datasets to natural large scale scene datasets.

We consider two scenarios here: a) a binary classification problem boat vs. tank, and b) a four class problem including boat, tank, bus and train. In each scenario, we train a SVM (using fc7 representation) from either natural scenes (selected from ImageNet) or ourDB and apply it to the other dataset. We also augment images from the two datasets and measure the accuracy on each individual dataset. We consider both off-the shelf features of the Alexnet (pre-trained over ImageNet) and fine-tuned features over our dataset.

\begin{table}
\renewcommand{\tabcolsep}{3mm}
\begin{center}{\footnotesize
\begin{tabular}{lc|c| c|c}
& \multicolumn{2}{l}{Without fine tuning}  & \multicolumn{2}{c} {With fine tuning}  \\
 \cline{2-5}
& \multicolumn{1}{l}{Natural} & ourDB & Natural & ourDB  \\
\cline{2-5}
\hline
\multicolumn{1}{l}{Natural} & 95 & 75  & 93 $\downarrow$ & 65 $\downarrow$\\
\multicolumn{1}{l}{ourDB} & 78 & 97  & 70 $\downarrow$ & 100 $\uparrow$\\
\end{tabular}}
\end{center}
\caption{Domain adaptation with boat vs. tank classification. }
\vspace{-5pt}
\label{tab:domainAll}
\end{table}

\begin{table}
\renewcommand{\tabcolsep}{1.5mm}
\renewcommand\arraystretch{1.4}
\begin{center}{\footnotesize
\begin{tabular}{ll|l| l|l}
& \multicolumn{2}{c}{Without fine tuning}  & \multicolumn{2}{c} {With fine tuning}  \\
 \cline{2-5}
& Natural & ourDB & Natural  & ourDB \\
\hline
\multicolumn{1}{c}{Natural [2000]} & 96.48 (0.5) & 55.6 (2.7)  & 95.56 (0.6) & 68.06 (2.0) \\
\multicolumn{1}{c}{ourDB [2000]} & 66.92 (3.2) & 96.90 (0.2)  & 65.22 (1.4) & 99.72 (0.1) \\
\hline
\multicolumn{1}{c}{ourDB [1000] + } &  &   &  &  \\
\multicolumn{1}{c}{Natural [1000]} & 94.42 (0.8) & 93.94 (0.4)  & 92.52 (0.2) & 98.70 (0.2) \\
\end{tabular}}
\end{center}
\caption{Domain adaptation over a 4-class problem (boat, tank, bus, and train). Numbers in parentheses are standard deviations.}
\label{tab:domainFourClass}
\vspace{-10pt}
\end{table}

\noindent \textbf{Augmenting data along all parameters:} 
Here we choose images along all parameters. Results in Table~\ref{tab:domainAll} show that training on each type of image, expectedly works the best on the same type of test image (95\% from ImageNet to ImageNet and 97\% from ourDB to ourDB). Cross application results in lower accuracy, but still above 50\% chance.

We find that fine tuning the Alexnet on our dataset boosts the performance on ourDB to 100\% with the cost of lowering the accuracy over the ImageNet. Doing so lessens other accuracies since the CNN features are tailored (and hence selective) to our images. The reason why performance is low when applying a trained model from our dataset to ImageNet is mainly because objects in these two datasets have different textures and statistics.

Table~\ref{tab:domainFourClass} shows domain adaptation results over 4 classes. Results confirm what we learned over 2 classes, although accuracies are lower here. We also found that similar to fine tuning, combining images from datasets hinders performance over each individual dataset due to contamination. 

\begin{figure}[t]
\begin{center}
   \includegraphics[width=1\linewidth]{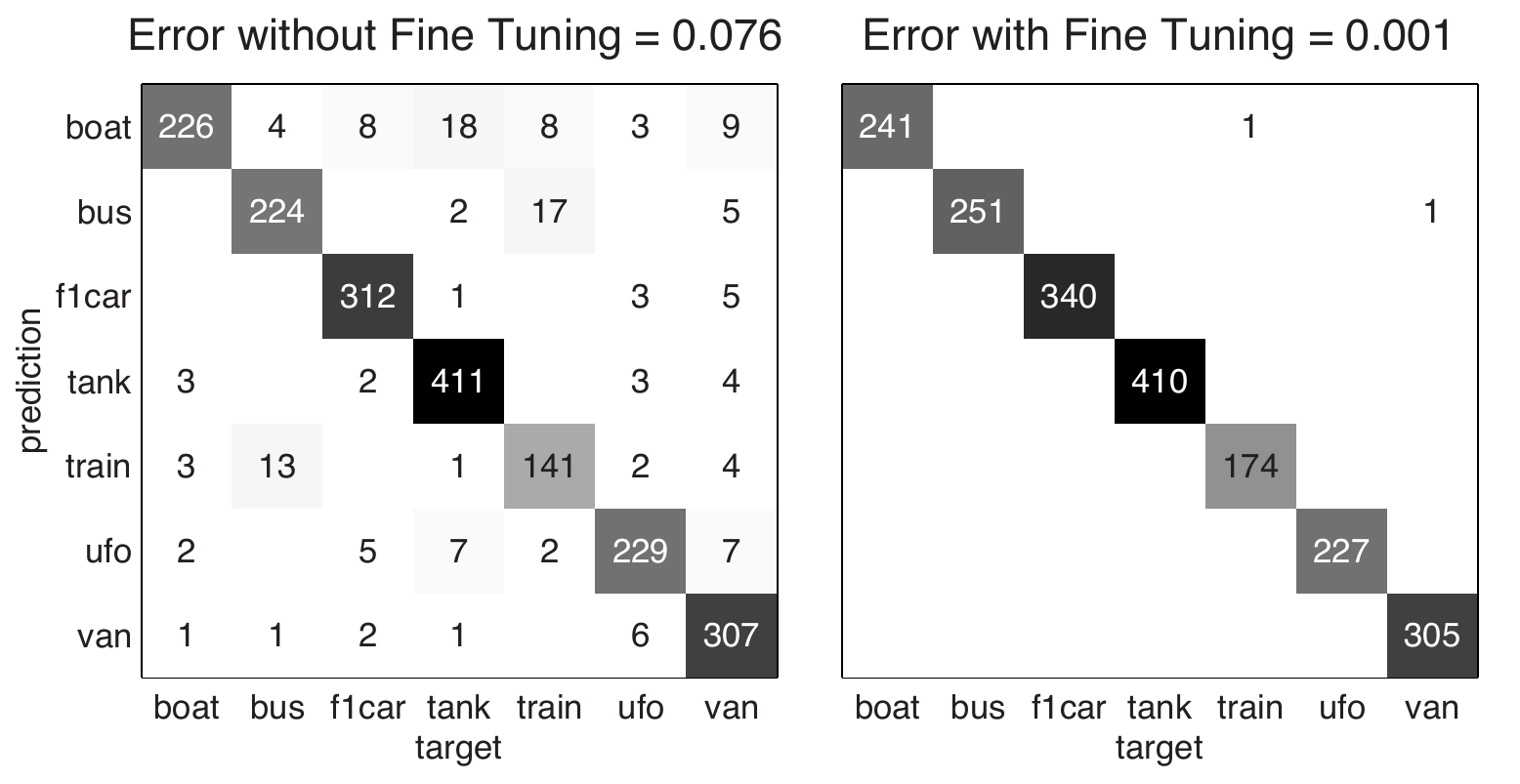}
\end{center}
\vspace{-10pt}
   \caption{Confusion matrices of Alexnet over 7 classes of our dataset without (left) and with fine tuning.}
\label{fig:confusion}
\vspace{-10pt}
\end{figure}

Performances over 2-class and 4-class problems were very high here (above 95\%). To further investigate accuracy of the Alexnet, we increase the number of classes to 7. As seen in the confusion matrices in Fig.~\ref{fig:confusion}, fine tuning the network increases the accuracy from 92.5\% to 99.9\% with only two mistakes\footnote{Please see the supplementary material for t-SNE visualization~\cite{van2008visualizing} of without- and with fine tuned fc7 and pool5 features.}.

\begin{figure}[t]
\begin{center}
   \includegraphics[width=1\linewidth]{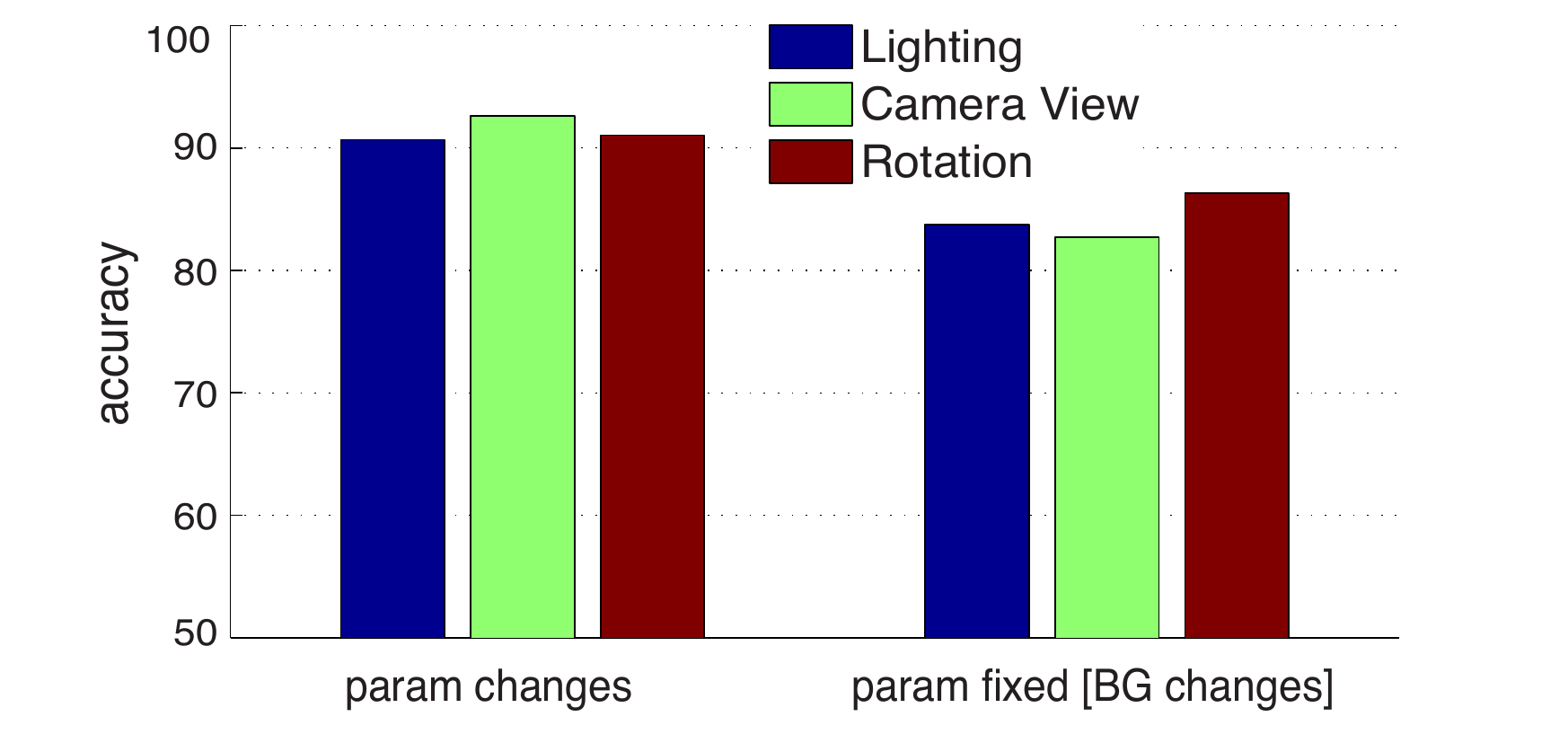}
\end{center}
\vspace{-8pt}
   \caption{Domain adaptation when a single parameter can change.}
\label{fig:domainSingle}
\vspace{-13pt}
\end{figure}

\noindent \textbf{Augmenting data along single parameters:} Here, we aim to see which parameter makes the most effect on domain-adaptation (from synthetic images to natural images.). We take two categories, boat and tank, as both synthetic and natural images are available for these two categories. While keeping all parameters fixed, we vary only one parameter to form a training set. Thus, we will come up with a customized training set in which only one dominant parameter is varying. Then, fc7 features are extracted for the training set and a linear SVM classifier is trained on these samples. The same features are extracted for the natural images and the learned model on synthetic samples is tested on them. For each parameter, we had 275 synthetic images for training and a fixed size set of 3000 images from ImageNet. To verify our findings, another experiment was designed in which all parameters were allowed to vary except a target one. 2000 samples were randomly selected which satisfied our constraints and a linear SVM was trained (using fc7). The parameter whose absence drops the accuracy more is considered to be more dominant on natural images. 5-fold cross validation accuracies are reported in Fig.~\ref{fig:domainSingle}.

As shown in the right side bars in chart in Fig.~\ref{fig:domainSingle}, it is clear that the camera-view is of higher importance as we gain the highest accuracy on the fixed natural test set. That does make sense, since in real world images, it is expected to see objects is different degree of elevations, and it is the dominant varying parameter in the wild. The rotation is the next important parameter as it gains the next top accuracy on natural images. Surprisingly, the lighting source ranked as the least effective parameter in our analysis. The right side bars in chart in Fig.~\ref{fig:domainSingle} verifies our findings so that absence of camera-view drops the recognition accuracy more than other two parameters.

\subsection{Analysis of parameter learning order}

In this section, we analyze how the order of knowledge delivery to CNNs influences parameter prediction. To do so, we first prepare 40K training and 10K validation sets while annotating them with rotation labels from four categories (boat, bus, tank and train). AlexNet baseline is fine tuned on this training set, hoping to start from a proper initialization point for optimization procedure. We set the learning rate for all weights to 0.001 except for those of fc8 where they are set to 0.01, and leave all other parameters to the default values.  Afterward, we prepare a new training set consisting of 40K images from the same four categories in the previous setting, except that they are annotated with camera view labels. 10K validation set is also prepared in the same way. Obtained weights from the previous step are loaded to the network and are treated as a promising initialization point for another fine tuning process over new data. The learning rate is set to 0.001 for all weights except the fc8, where they are set to 0.01. All other parameters are left to have the default values. 

Next, we evaluate the performance on camera view and rotation prediction using pool5 layer representation. As fine tuning with low learning rate slightly changes weights within the network, we are interested to see which order of changes in weights (before fully connected layers) gives the superior performance in our desired task. 
To hunt what we are looking for, order of prepared datasets is reversed and delivered to the network in the opposite way, i.e. first camera and then rotation. We denote aforesaid orderings as follows: 1) rotation-camera, and 2) camera-rotation for simpler reference. In the evaluation phase, 2000 samples are randomly selected from four categories, and pool5 features for them are extracted. After mean subtraction and dimensionality reduction, accuracies of 5-fold cross validation are reported (See Table.~\ref{tab:order}).

Counter-intuitively, we found that order of data delivery is so important to the network such that when the network sees the samples with rotation labels prior to camera labels, it ostensibly performs better in parameter prediction. From the results in Table.~\ref{tab:order}, we can see when the network is firstly fine tuned on rotation, the second stage, i.e. fine tuning on camera labels, does not damage the weights for rotation prediction. In contrast, when the camera labels are seen by the network before rotation labels, performance of rotation prediction is expectedly becoming better than the previous ordering, however this boost causes dramatic degradation in camera prediction.

\begin{table}
\begin{center}
\renewcommand\arraystretch{1.1}
\begin{tabular}{l|cc}
Task & 1  [rotation-camera]  & 2  [camera-rotation]  \\
\hline
Camera  & 89.20\% (1.47)  & 77.05\% (1.18) \\
Rotation  & 93.75\%(1.66)  & 95.30\% (1.00) \\
\end{tabular}
\end{center}
\caption{Influence of data delivery order on parameter prediction.}
\label{tab:order}
\vspace{-10pt}
\end{table}

As we found in our previous experiments, camera view variation is a more ill-structured parameter to predict. When the network sees the camera labels in the second stage, the adapted weights are more biased towards learning this parameter, while the shiny point is that this bias does also try to keep the pre-seen knowledge for rotation unchanged. Hence, we can conclude that when there is the option for stage-wise training, it would be better to sort the parameters according to their complexities and feed them to the network following simple to complex order. This way, the last steps are devoted to manage the difficulties in complex parameters, while imposing less damage to weights adapted for simpler parameters.

\section{Discussion and conclusion}

We challenged the use of uncontrolled natural images in guiding that object recognition progress and introduced a large scale controlled object dataset of over 20M images with rich variety of parameters that can be useful in the field. By choosing slices through our dataset, we were able to systematically study the invariance properties and generalization power of CNNs by independently varying the choice of object instances, viewpoints, lighting conditions, or backgrounds between training and test sets. Progressively extending these results on increasingly
larger subsets of our dataset may help gain new insights on how the algorithms can be modified to show
greater invariance and generalization capabilities. In what follows, we summarize the lessons we learn from our empirical investigation of the Alexnet baseline on synthetic and natural images.

\vspace{3pt}
\noindent \textit{i)} Representation learned in pool5 layer is selective to parameters (it is possible to readout parameters) while fc7 layer is not. Both of these layers contain object category information (fc7 is more selective). It would be interesting to explore how selectivity of fc7 to both object and parameters can be increased simultaneously.


\vspace{3pt}
\noindent \textit{ii)} The knowledge obtained from some parameters is easier to be transferred to unseen object categories. In particular, we saw that illumination possesses the simplest knowledge, whereas rotation and camera-view parameters are more difficult. We also found that 3D properties of the object shape play a critical role in knowledge transfer. The higher variability in shape, the less knowledge transfer on unseen categories.



\vspace{3pt}
\noindent \textit{iii)} Results of our sampling strategy analysis revealed the importance of instance level variety compared to that of parameter level. In particular, we found that random sampling strategy leads to better generalization since more instance level variations can be included in the dataset. 



\vspace{3pt}
\noindent \textit{iv)} Results of data augmentation shows that simply adding instances from two classes does not improve accuracy mainly because objects in these two datasets have different textures and statistics. However, we found that there is generalization from one dataset to the other as cross application of one dataset to the other results in above chance accuracy. It would be interesting to learn functions for domain adaptation from our images to natural real world scenes such as those in the ImageNet dataset.


\vspace{3pt}
\noindent \textit{v)} A large scale synthetic object database, such as the one presented here, could be used as a diagnosis tool to infer along which dimensions a large scale wild dataset varies the most and how wild datasets offer information regarding invariance to parameters.

\vspace{3pt}
\noindent \textit{vi)} Last but not the least, we found that when there is the option to perform stage-wise training, it would be advantageous to feed the network with data that has been sorted according to complexities of different dimensions. This can lead us to train CNNs layer-wise for learning different invariances in different layers.

Currently, deep learning models sacrifice invariance in favor of higher object category prediction accuracy. It would be best if we can achieve both at the same time (e.g., it might be needed in some applications). It might be possible to organize the feature manifolds in the the early layers in such a way to preserve information about object parameters as well as category information. Two ways to explore this include feature embedding through loss regularization or adding camera parameters to the categorization loss. The idea would be knowing the camera parameters may help object categorization. A recent study~\cite{elhoseiny2015convolutional} have investigated this idea by proposing a convolutional network for joint prediction of object category and pose information.

In summary, we answered some questions regarding CNNs and datasets, and discussed future large-scale applications of our dataset, which is freely shared and available.

\textbf{Acknowledgements:} We wish to thank NVIDIA for their generous
donation of GPUs used in this study.

{\small
\bibliographystyle{ieee}
\bibliography{biblio}
}

\newpage

\noindent \textbf{---------------------------  Appendix ------------------------------ }

We use t-SNE dimensionality reduction method~\cite{van2008visualizing} to visualize the learned representations 

\vspace{5pt}
\noindent \textbf{Experiment I: category prediction}
In this experiment, we randomly select 2K samples from 7 categories (boat, bus, f1car, tank, train, ufo, and van) and feed them to a pre-trained CNN model, specifically Alexnet. Having fc7 and pool5 representations of selected samples ready, we use the t-SNE algorithm to reduce their dimensionality to 2D. 

In addition, 20K images are randomly selected from all 7 categories and the network is fine-tuned on the provided data for object categorization. The same procedure is carried out on the fine-tuned (FT) network. Fig.~\ref{fig:category} depicts the results.

Our results in Fig.~\ref{fig:category} show that fc7 representation works remarkably well at recognizing object level categories as they are mutually linearly separable after fine-tuning the network. Furthermore, pool5 representation does not contain discriminative information between object categories compared to fc7. This result is in alignment with Bakry et al.,~\cite{bakry2015digging}. Fig.~\ref{fig:category} also demonstrates the effect of fine-tuning on feature spaces. The distributions of samples for different categories tend to become very compact and concentrated after fine-tuning. Notice that fine-tuning does not add more discriminative power to the pool5 representation. 



\begin{figure*}[t]
\begin{center}
\includegraphics[width=1\linewidth]{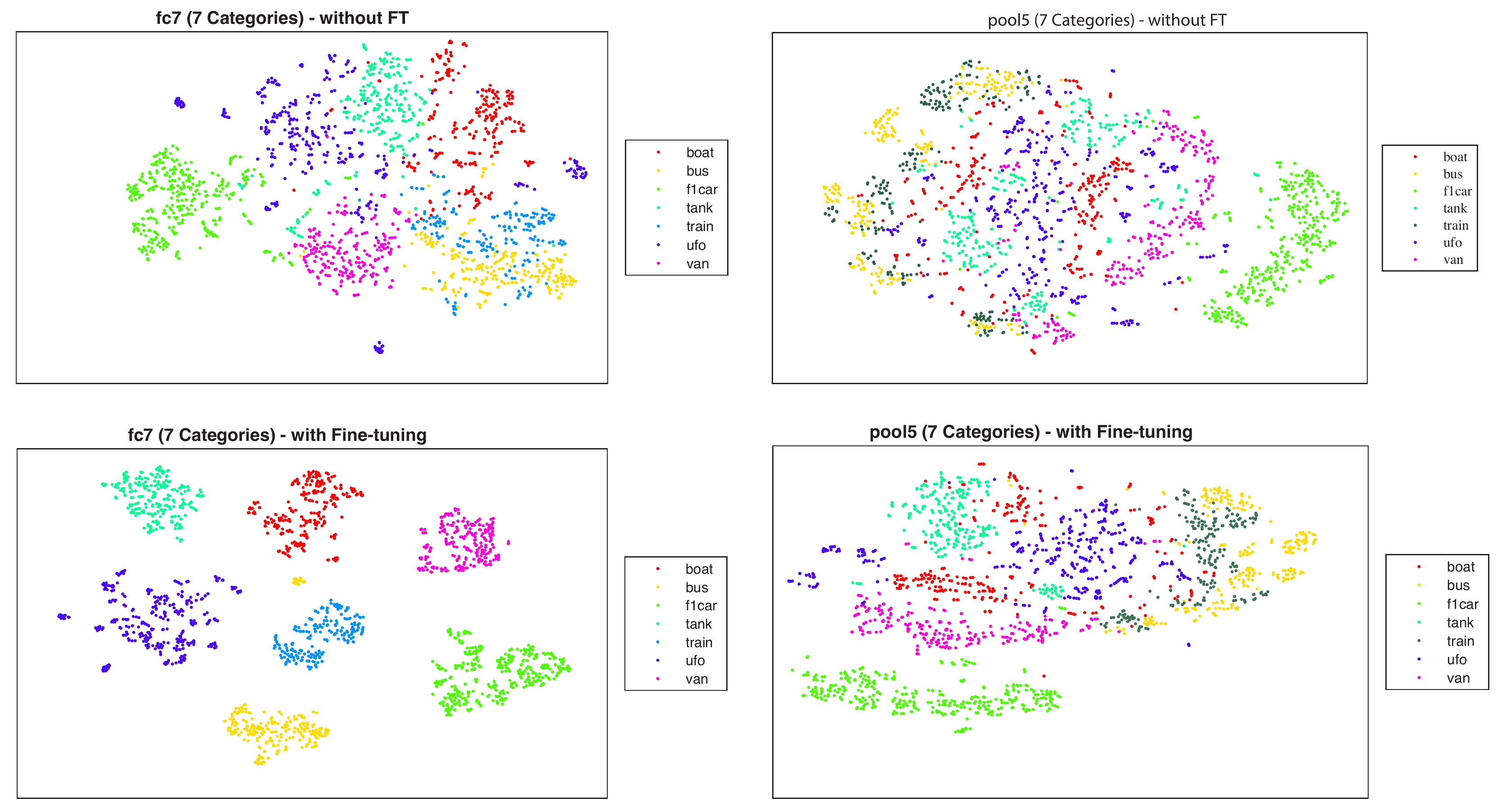}
\end{center}
\caption{t-SNE representation for category prediction using fc7 and pool5 layers with and without fine-tuning.}
\label{fig:category}
\end{figure*}

\begin{figure*}[t]
\begin{center}
\includegraphics[width=1\linewidth]{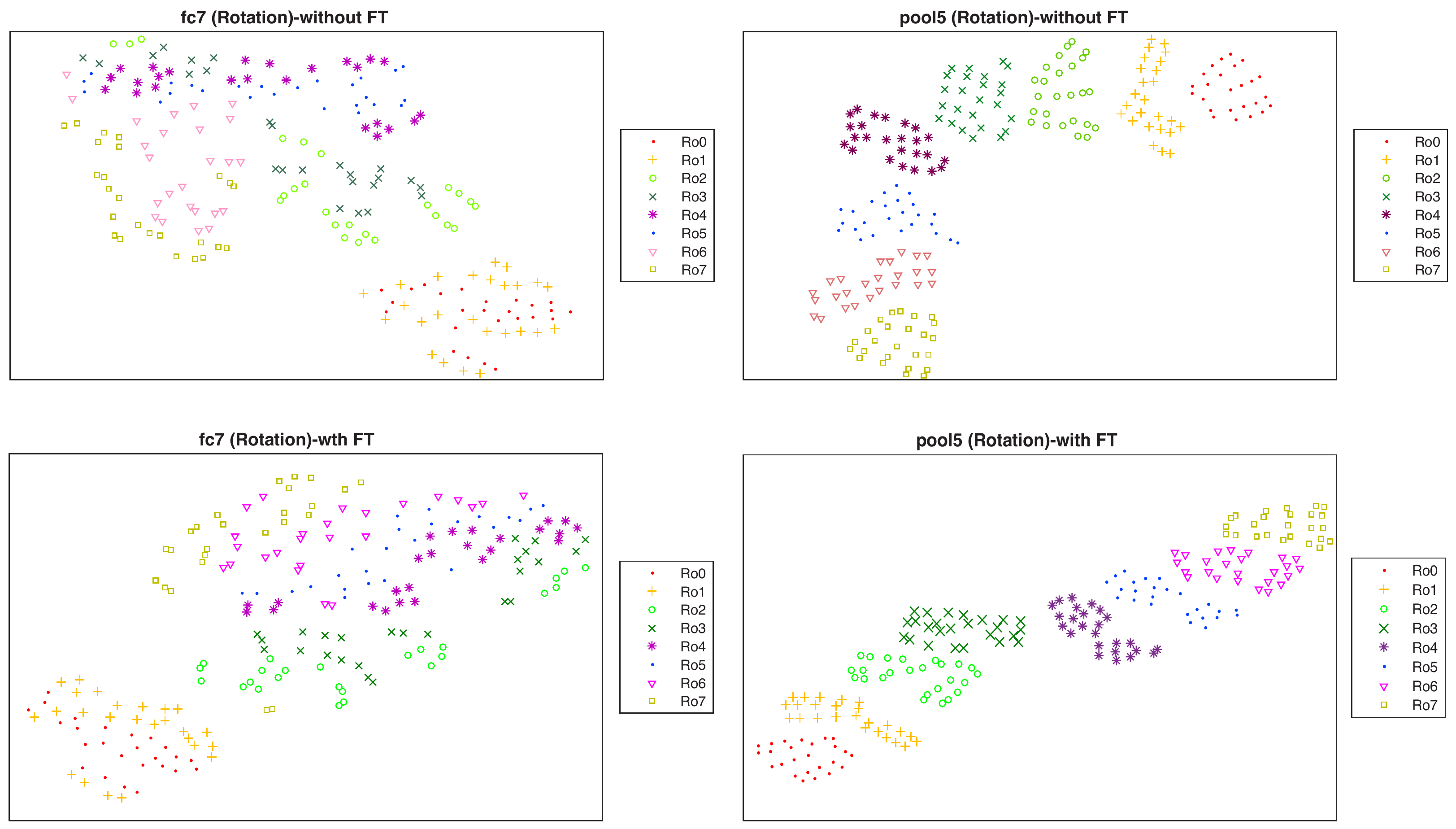}
\end{center}
\caption{t-SNE representation for lighting prediction using fc7 and pool5 layers with and without fine-tuning.}
\label{fig:rotation}
\end{figure*}

\vspace{5pt}
\noindent \textbf{Experiment II: rotation prediction}
This experiment makes effort to highlight the power of pool5 layer in representing image variations and discriminating among them. As we discussed in the main paper, our analyses show that pool5 representation gives superior performance for parameter prediction. To confirm this statement, we select 200 samples from the boat category (and instance number 01) while rotation, camera, and lighting parameters are changing. We then label the samples with their rotation values and feed them to the pre-trained Alexnet model. The dimensionality of fc7 and pool5 representations are reduced to 2D using tSNE. The same procedure is carried out using the fine-tuned network to obtain the fc7 and pool5 representations. Results are illustrated in Fig.~\ref{fig:rotation}.

It can be seen that fc7 representation is not (fully) capable of discriminating the rotation values, both with and without fine-tuning. The representation by the pool5 layer, in contrast, confirms our findings that pool5 contains information selective to parameters. Samples from 8 different rotation values are perfectly and mutually linearly separable from each other. Fine-tuning tries to improve the discriminability through some sort of transformation.


\begin{figure*}[t]
\begin{center}
\includegraphics[width=1\linewidth]{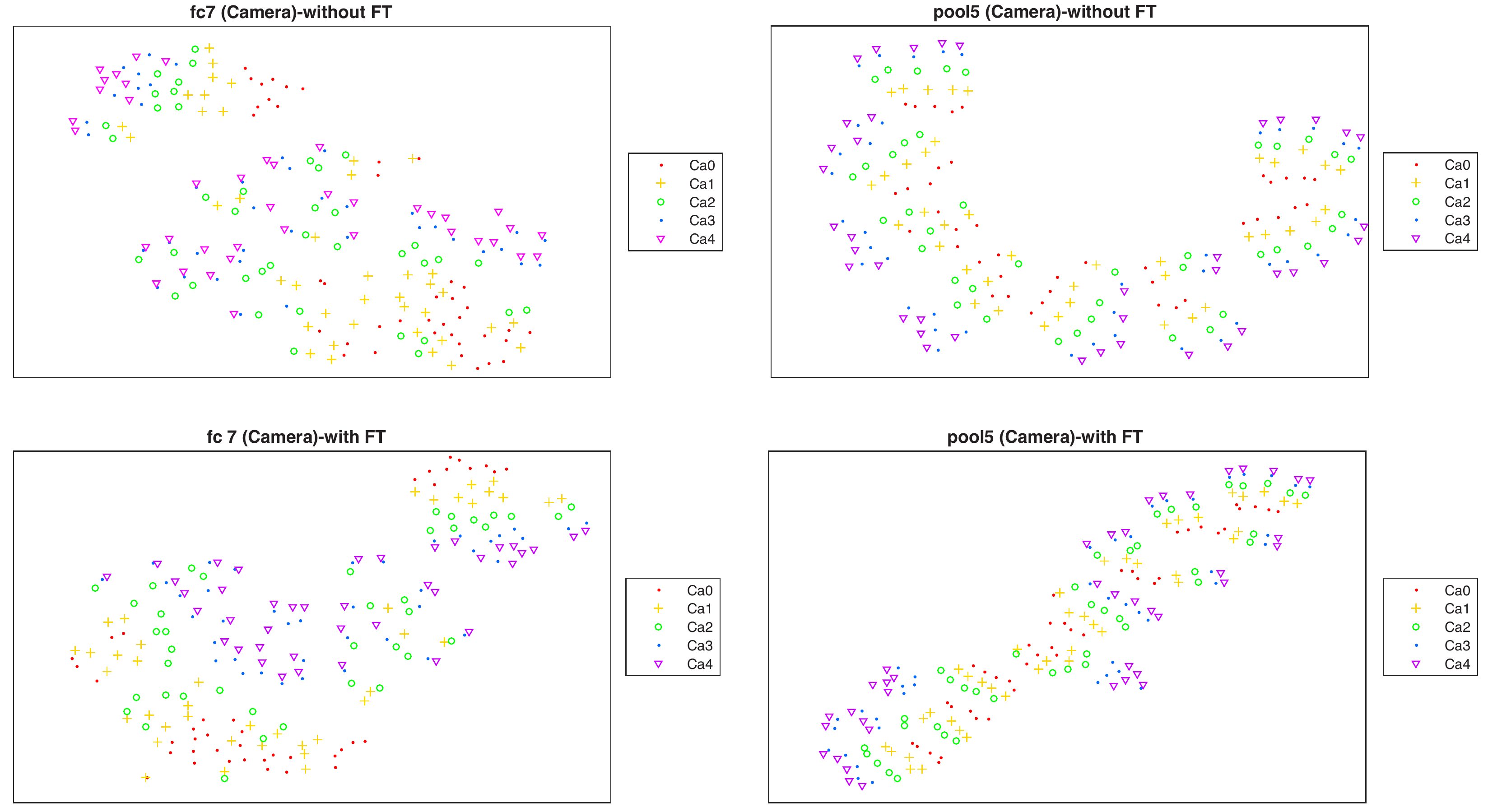}
\end{center}
\caption{t-SNE representation for camera prediction using fc7 and pool5 layers with and without fine-tuning.}
\label{fig:camera}
\end{figure*}

\vspace{5pt}
\noindent \textbf{Experiment III: camera prediction}
With our success in visualizing the power of pool5 layer in capturing rotation variations, in this experiment we aim to see whether the same judgment is valid for camera prediction. As in the previous experiment, we select 200 samples from the boat category (instance number 01) and label them according to their camera parameter value. 2D feature spaces derived from fc7 and pool5 representations using pre-trained and fine-tuned Alexnet are depicted in Fig.~\ref{fig:camera}.

As before, fc7 representation does not offer useful information regarding separating samples with different camera parameters, both in pre-trained and fine-tuned cases. We observe quite the opposite using the pool5 layer representation. Without fine-tuning the network, we can observe 8 clusters in Fig.~\ref{fig:camera} (see the up-right panel), each one corresponding to one rotation. For each rotation angle, the representation is surprisingly capable of discriminating different values of camera parameters in five classes (we only use five values for camera parameter here).



\begin{figure*}[t]
\begin{center}
\includegraphics[width=1\linewidth]{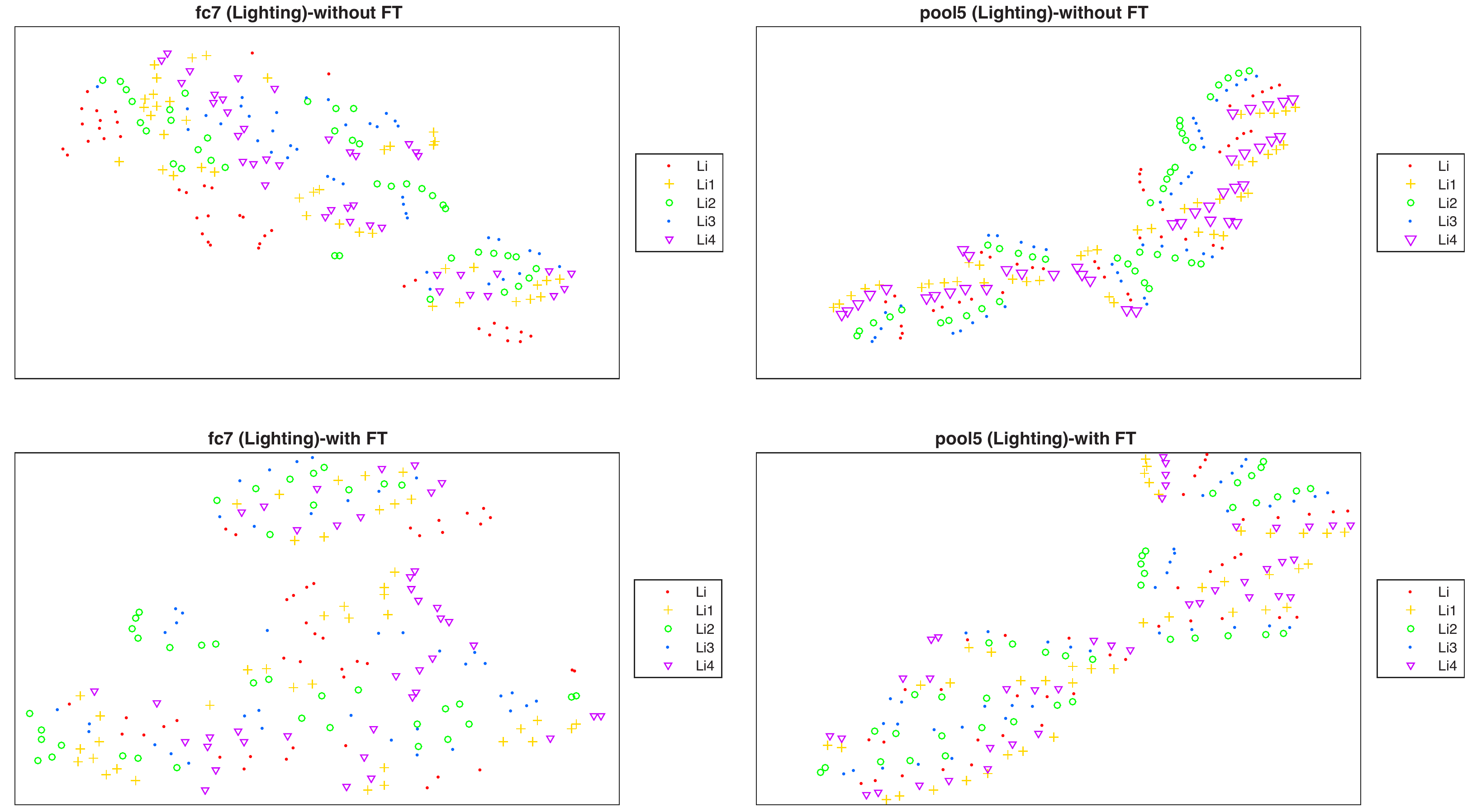}
\end{center}
\caption{t-SNE representation for lighting prediction using fc7 and pool5 layers with and without fine-tuning.}
\label{fig:lighting}
\end{figure*}

\vspace{5pt}
\noindent \textbf{Experiment IV: lighting prediction} Scrutinizing the behavior of fc7 and pool5 layers should be interesting for lighting prediction as well. Therefore, we follow the previous experiments except that here samples are labeled according to the lighting parameter values. Fig.~\ref{fig:lighting} shows the results for four different cases. 

Skipping the poor representation by fc7 layer, pool5 layer again generates reasonable representation which is able to discriminate between different lighting conditions. Eight clusters are observable, each one corresponding to one rotation angle. In each cluster, samples with different lighting parameters are discriminant which again supports our previous statement regarding the capability of the pool5 layer in parameter prediction. 



\begin{figure*}[t]
\begin{center}
\includegraphics[width=1\linewidth]{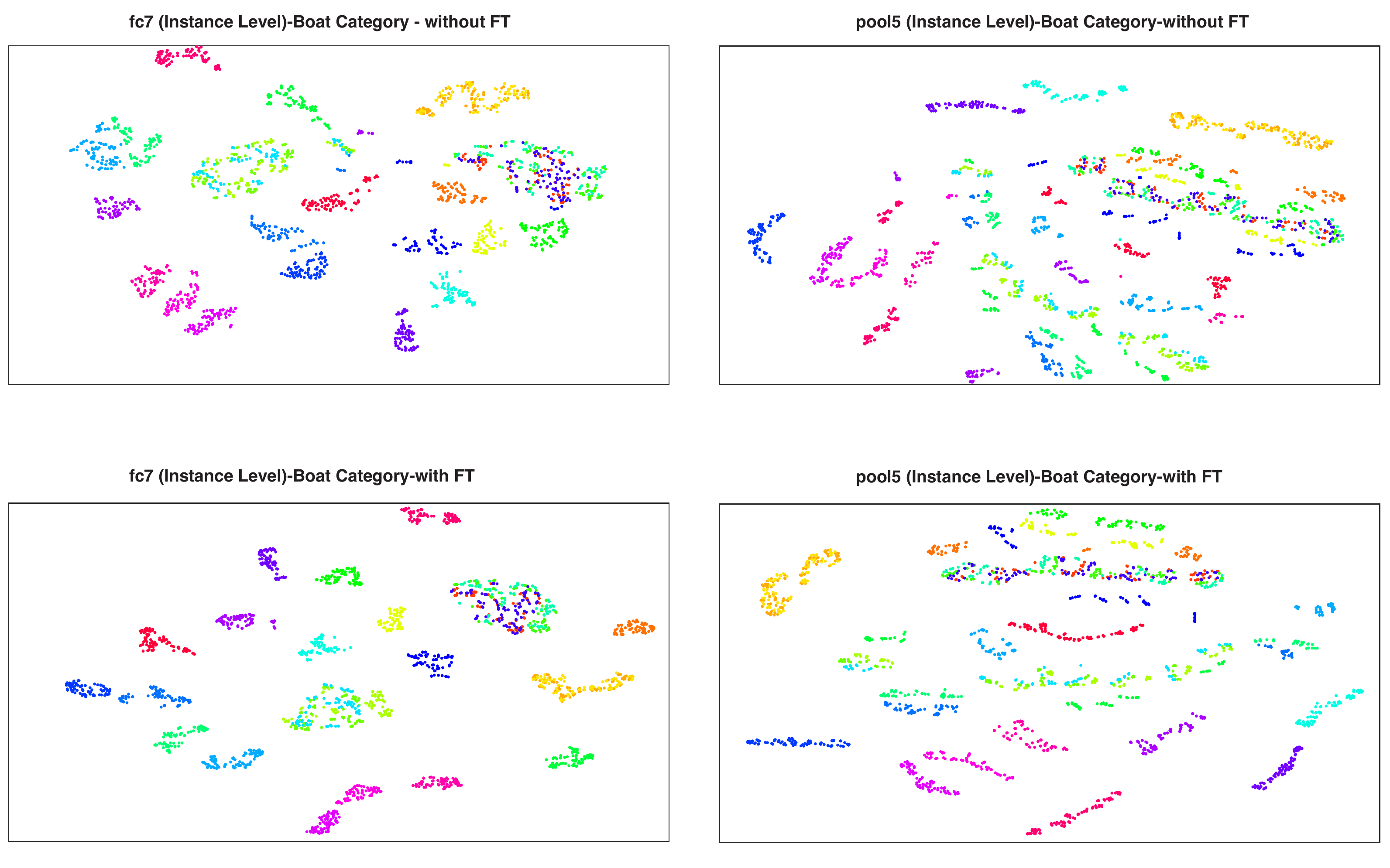}
\end{center}
\caption{t-SNE representation for instance prediction using fc7 and pool5 layers with and without fine-tuning.}
\label{fig:instance}
\end{figure*}

\vspace{5pt}
\noindent \textbf{Experiment V: instance prediction} In the last experiment, we aim to inspect the capacity of fc7 and pool5 layers of CNNs for instance prediction. We randomly choose 2K samples from the boat category. The samples are passed through the network up to pool5 and fc7 layers. The obtained representations are visualized after dimensionality reduction using the tSNE. The same procedure is repeated with the fine-tuned network. Fig.~\ref{fig:instance} show the results. 

The fc7 representation, without fine-tuning, is remarkably capable to separate samples from different instances. Fine-tuning the network dramatically boosts this discrimination power by making clusters more compact. A representation is invariant to varying parameters if it ignores variations and treats samples with different parameters equally, i.e., it  makes the representations of similar samples as close as possible in the feature space. This is exactly what we see in the in the representation space provided by fc7. 

Despite the reasonable parameter separability, the pool5 layer does not force different instances to be clustered. This is the place where difference between pool5 and fc7 layers can be seen in practice. This result indicates that the fc7 layer seeks to produce invariant representations (by collapsing manifolds), while the pool5 layer tries to preserve manifolds as much as possible.

\end{document}